\documentclass[a4paper, 12pt]{extarticle}

\usepackage{graphicx}
\usepackage[colorinlistoftodos]{todonotes}
\usepackage[margin=1.0in]{geometry}
\usepackage{enumitem}
\usepackage{indentfirst}
\usepackage{makecell}
\usepackage[tableposition=top]{caption}
\usepackage{listings}
\usepackage{url}
\usepackage[round]{natbib}
\usepackage[colorlinks=true,citecolor=blue,linkcolor=blue,urlcolor=blue]{hyperref}
\usepackage{pgfgantt}
\usepackage{framed}
\usepackage{adjustbox}
\usepackage{lipsum}
\usepackage[justification=centering]{caption}
\usepackage{setspace}

\usepackage{mdframed}
\title{Using Textual Summaries to Describe a Set of Products}
\author{Kittipitch Kuptavanich \\
{kittipitch.kuptavanich@abdn.ac.uk}}
\date{}

\begin{document}

\maketitle
\section{Introduction}
\label{sec:introduction}

% Suppose a customer is faced with a choice between what Tanner and Raymond call \textit{``high-involvement products''}, that is, products that are inherently complex, which carry a high risk of failing, or have a high price tag \cite{Tanner2011}. The authors observe that, given a large set of products to chose from, most consumers would not be interested in examining all of them.

Tanner and Raymond \cite{Tanner2011} observe that, when consumers are faced with a set of products to choose from, most would not be interested in examining them exhaustively. Instead, they start with \textit{``information searching''}. During information searching, customers acquire a basic overview of the product set as a whole \cite{Tanner2011}: for instance, how many subtypes there are in the set, what features to expect in a typical product of this type, what is the price range, and so on. Based on this information, the consumer can then develop \textit{``evaluation criteria''} to help narrowing down choices.

%In a decision making process, a recommender system can also narrow down the choices into sets of different sizes. Regardless of the size of the recommended set, a good overview in forms of description of the set can aid as an explanation benefiting the consumer in various ways, e.g. trust, effectiveness, persuasiveness, efficiency and satisfaction \cite{Tintarev2007} while engaging in Tanner/Raymond-style information searching upon the set.

% We believe that Tanner and Raymond's ideas are relevant for Recommender Systems (RS) in a number of ways: An overview of a set of products may help the consumer in various ways, enhancing trust, effectiveness, persuasiveness, efficiency and satisfaction \cite{Tintarev2007}; additionally, where a Recommender System has insufficient information to confidently recommend one item (e.g., during a ``cold start'') \cite{Schein:2002:MMC:564376.564421}, then a Recommender System may opt to present a larger set of items, necessitating Tanner/Raymond-style information searching.

We believe that Tanner and Raymond's ideas are relevant to Recommender Systems (RS) in a number of ways. An overview of a set of products can enhance an RS, by increasing trust, effectiveness, persuasiveness, efficiency and satisfaction \cite{Tintarev2007} when consumers engage in Tanner/Raymond-style information searching upon the set.
Additionally, where an RS has insufficient information to confidently recommend one item (e.g., during a ``cold start'') \cite{Schein:2002:MMC:564376.564421}, then a Recommender System may opt to present a larger set of items, necessitating Tanner/Raymond-style information searching.

My PhD project explores the hypothesis that a consumer's decision making can be aided by an approach inspired by Shneiderman's Visual Information Seeking mantra \cite{545307}. This mantra is often summarised by the slogan, ``{\em Overview first}, zoom and filter, then details-on-demand''. We focus here on the first part of Shneiderman's slogan (italicised), namely that it is beneficial for a reader to be exposed to an overview before diving into specifics.

% My PhD project explores the hypothesis that information searching can be aided by an approach inspired by Shneiderman's Visual Information Seeking mantra \cite{545307}. This mantra is often summarised by the slogan, ``{\em Overview first}, zoom and filter, then details-on-demand''. We focus here on the first part of Shneiderman's slogan (which we italicised), namely that it is beneficial for a reader to be exposed to an overview before diving in to specifics.

Textual overviews of large sets of consumer products are not a new idea, of course. However, we observe that such overviews are written by hand \cite{which,consumerreports}.
% and this limits their applicability in Recommender Systems.
% , and also in scenarios where a consumer searches for products with particular keywords (e.g. on an e-commerce website).
While static manual summaries are typically provided for only top level categories, dynamic computer generated summaries can  be provided for any set of products, for example those filtered by users with their criteria of interest. Our research hypotheses is that useful overviews of consumer product sets can be generated automatically by computer, using Natural Language Generation (NLG) techniques. If confirmed, this would be a potentially important finding for both the Recommender Systems and the Information Retrieval community.

To illustrate my work so far, I will present description of our automatically generated summaries inspired by hand-written summaries, followed by an evaluation with human users. Section~\ref{sec:related work} summarises relevant work in NLG and automated summarisation.  Section~\ref{sec:algorithm} describes our automatically generated summaries based on our observation over handwritten summaries. Section~\ref{sec:experiment} presents an experiment in which we tried to assess whether the summaries generated by our algorithm helps users ``understand'' the information in two product sets, and whether it helps them make a well-founded choice. Section~\ref{sec:discussion} discusses how our findings point the way towards even more useful computer-generated product summaries.

\section{Related Work}
\label{sec:related work}

\subsection{Product Comparison Interfaces}
Many websites support product comparison, mostly in a tabular format, where users can select 2 or more products and features to compare side-by-side \cite{gadgetsnow,saveonlaptops}. Consumers are often able to apply feature based filters to narrow down the set of products \cite{pricespy,opodo,uswitch}. Product information is often presented as a table of specifications with no accompanying textual summary of the items presented in the table.

\subsection{Natural Language Generation} % Shorten!
% KvD Added
Within Computational Linguistics, the automatic generation of text from non-textual input is addressed in the research area of Natural Language Generation (NLG) \cite{Reiter:2000:BNL:331955,2017arXiv170309902G}.

One area of NLG that is potentially relevant here is Referring Expressions Generation, where the aim is to identify a referent for a hearer (i.e., so the reader knows what it is). As in our case, the referent may be a set, for instance as when an NLG system generates ``the blue sofas'' to enable a reader to know what sofas the writer has in mind \cite{Deemter2002}. Unlike most previous work, however, our aim is not to allow the reader to know \textit{which consumer items} the system is talking about: the purpose, rather, is to give the reader insight in the broad composition of the set (e.g., so s/he knows what the main commonalities and difference across the set are).\cite{2014Kutlak}

Given what we observed about the prevalence of quantified statements in product set surveys (Section~\ref{sec:algorithm}), another potentially area of NLG is where NLG algorithms extract trends and patterns from data. Perhaps the most sophisticated example is Narratives for Tableau \cite{narratives_for_tableau}, a commercial product that generates text from analyzing data associated with user-selected areas of a chart made with Tableau. The extension produces description of the data such as  \textit{``Sales and profit ratio moved in opposite directions from January 2011 to December 2014.''} However, the extension focuses on time-series data, and it is difficult to see how Tableau's techniques could be used for giving insight into a static set of product. A similar example is Automatic Statistician \cite{automaticstatistician}, which also does time-series data to text generation using statistical methods. To our knowledge up to this point, there is no NLG system that describes set of items.

\subsection{Explanation of Recommender System}
Item description accompanying recommended set can be designed to benefit readers with various aims in mind. \cite{Tintarev2007} classified them as Transparency (Tra) - Explain how the system works, Scrutability (Scr) - Allow users to tell the system it is wrong, Trust - Increase users’ confidence in the system, Effectiveness (Efk) - Help users make good decisions, Persuasiveness (Pers) - Convince users to try or buy, Efficiency (Efc) - Help users make decisions faster and Satisfaction (Sat) - Increase the ease of usability or enjoyment. A dynamically generated description of recommended items, when used as an explanation, could potentially be an enhancement to a Recommender System in multiple aspects.

\section{Automatic summarisation of product sets}
\label{sec:algorithm}

%\section{Analysing hand-written summaries} % Shorten. Focus on findings.
%\label{sec:corpus analysis}

Many websites contain hand-written reviews of product sets.\cite{which,consumerreports,saveonlaptops,gadgetsnow,uswitch,dpreview}. Here we list the lessons we observed that informed the algorithm we used to generate our first summary. As expected, most reviews started with an {\em introductory paragraph} that surveyed the product set as a whole and sketched the shape of the price curve of the product set. Subsequently, many reviews contained sentences that {\em quantify} over the product set, using patterns like \textit{``Most [products] are/have...''}, \textit{``Many [products] have...''}. Reviews also tended to say which features they should {\em pay attention to}.

%A good example of the above can be found in\cite{which_tv_overview}.

%\textit{``...here are some features to look out for:
%    \begin{itemize}
%        \item \textbf{Full-HD display}: To get the best possible picture quality, go for a 1080p, or Full-HD, TV. An HD-Ready set will let you view HD content but it typically won't be as good as a Full-HD model. ...%New 4K TVs are available, but they're mainly in sizes of 40-inches and over.
%        \item \textbf{Built-in HD tuner}: ...%You'll need HD content in order to enjoy your high-definition display to the full. Go for a 32-inch TV with a built-in Freeview HD or Freesat HD tuner, allowing you to get subscription-free digital TV channels. Some TVs even have both.
 %       \item \textbf{Smart-TV}: ...%Most new TVs from the big manufacturers come with smart-TV features, allowing you to watch catch-up and on-demand services such as BBC iPlayer and Netflix. ... %But cheaper TVs from smaller brands may miss out. Head over to our What is smart TV? guide to find out if this is an important feature for you.''
 %   \end{itemize}}

As a result of our observation, we implemented an NLG system using simple template-based approach with jinja2 template engine \cite{jinja2}. From the 4 stages of data-to-text NLG architecture \cite{Reiter2007}, the majority of our work belongs to the content selection process of the \textit{Document Planning} stage. That generated summaries consisted of 3 parts namely, an introductory paragraph, followed by a collective description of products, and important feature highlights each in their own paragraph as shown below.
% \begin{lstlisting}[caption={An Example of a Summary Produced}, captionpos=b, label={lst:example_summary}]
\bigskip

\begin{mdframed}
\indent \textit{For 32 inch TVs, the price of most products (340 out of 363 models) falls in the range of 70-580 pounds with a median price of about 255 pounds.}

\textit{Most 32 inch TVs have following features: 16:9 aspect ratio, LED backlight, LCD display technology, HDMI, Flat panel design, analogue TV tuner, and digital TV tuner}

\textit{The features that have a strong impact on the price of 32 inch TVs are: number of hdmi inputs, release year, brightness, resolution, hd ready 1080p (full hd), smart TV, and annual energy consumption}
\end{mdframed}

\bigskip

{\bf Introductory Paragraph: Shape of the price curve.}
In this part of the summary, we simplified the task by focusing on providing information  describing the shape of the price curve of the set of product only. The price shape of the product are reported as a range of price, without the outliers, rounded to the nearest 5 (the outliers were identified using median absolute deviation: MAD).

{\bf Collective description of products: Common Features.}
In this part of the summary, we chose to report the most common features across the set of products.
From the set of products we identified the 7 top-most common features then reported it.

{\bf Highlighting Important Features.}
Since the items we are trying to report are consumer products, we theorised that the features that display strong effects on the price are the important ones that the consumers should focus on.

So for each features, we find average price for each subset.
For example, if an interested feature of a TV is `resolution', then the subgroup are 720p, 1080p, and 4K. We then find the average price of each subgroup. Then we again find the SD of the 3 average prices. The higher the SD, the greater effect on the price that feature has. We then ranked the top most 7 features using this comparison and generated a report.

\section{Evaluation Experiment}
\label{sec:experiment}
NLG algorithms have traditionally been evaluated in a number of ways \cite{TypesOfNLGEvaluation}. Given that our present aim is to produce texts that are useful to a reader (rather than texts that mimic a speaker), metric-based evaluations, such as BLEU for instance \cite{Papineni:2002:BMA:1073083.1073135}, are not very suitable. It therefore seemed to us that the most apt methods in our case are evaluation by means of human judgment and task-based evaluation. We decided to conduct a simple version of each of these evaluation methods, to see what they might teach us about our algorithm.

We wanted to find out whether automatically generated summaries of large sets of products can help customers make an informed decision about what product to buy; to do this, we focused on summaries generated by the above algorithm, which focuses on listing common product features and on listing features that have a strong effect on price. We wanted to know two things in particular: first, we wanted to find out how useful readers {\em believe} our summaries to be for selecting the products of interest to them; second, we wanted to make a first attempt at finding out whether our summaries actually {\em helped} participants to quickly identify those products that they are interested in.

To answer the first question (about participants' subjective appreciation of the summaries) we asked participants to answer four Likert-style questions that address the {\em perceived} usefulness of the summaries (which were mapped to different aims of explanation); to answer the second question (about the actual usefulness of the summaries) we asked participants to make a quick choice (``speeded choice'') from among all the products in the set after reading our summaries, and we compared these speeded choices with the choices that they would later make at their leisure (``gold-standard choice''); the smaller the difference between speeded choice (facilitated by our summaries) and gold-standard choice (reflecting participants' real preference), the more useful we considered the summaries to be. Our experiment thus consisted of Laboratory Human Rating and laboratory task-based evaluation. This idea will be explained and discussed in the following sections.

\subsection{Method}

\subsubsection{Materials}

We used two different product information databases:
\begin{itemize}
    \item{database of TVs containing 363 products (rows) with 100 features (columns)}
    \item{database of photo cameras containing 610 products (rows) with 71 features (columns) }
\end{itemize}
Both databases contained data scraped from \cite{pricespy} during June 2017. They were presented as spreadsheets in MS Excel format.

We chose large databases to make the task of ``understanding'' their content and selecting the most interesting items in them particularly challenging.

For the baseline group, we used an ultra-short summary that was designed to be truthful without being particularly helpful, as shown below.

Example Summary in Baseline Condition:
\\{\em For DSLR Cameras, the price of most products (550 out of 610 models) falls in the range of 10-1850 pounds with a median price of about 525 pounds.}

For the experimental group, we used the summary produced by our algorithm.

Example Summary in Full Summary Condition:
\\{\em For 32 inch TVs, the price of most products (340 out of 363 models) falls in the range of 70-580 pounds with a median price of about 255 pounds. Most 32 inch TVs have following features: (followed by 7 features). The features that have a strong impact on the price of 32 inch TVs are: (followed by 7 features)}

\subsubsection{Participants}

Participants were 16 graduate students in Computing Science and Chemistry Department of (ANOMYMISED) recruited through the departments' internal student mailing lists.

\subsubsection{Design and Procedure}

In total, there were thus 4 conditions:
\begin{itemize}
\item Condition A: Camera + Baseline Summary, TV + Full Summary
\item Condition B: TV + Full Summary, Camera + Baseline Summary
\item Condition C: Camera + Full Summary, TV + Baseline Summary
\item Condition D: TV + Baseline Summary, Camera + Full Summary
\end{itemize}

To find out about participants' subjective appreciation, we asked each participant a number of Likert-style questions about the usefulness of the summaries that they had seen. We used 5 Likert values, from 1 (I do not agree at all) to 5 (I fully agree). % KvD I made this up, so please replace this by the correct explanation.
\begin{itemize}
    \item ``Does the summary help you get a rough picture of the products in this category?'' [Q1] -- Trust
    \item ``Does the summary help you select the products from the table faster?'' [Q2] -- Efficiency
    \item ``Does the summary help you select the products from the table with more confidence?'' [Q3] -- Effectiveness
    \item ``Do you find the summary useful?'' [Q4] -- Satisfaction
\end{itemize}
To find out about the effectiveness of the summaries, We asked each participant to list their top-5 products (e.g., their top-5 TVs). Participants were explicitly told that the order of the items in their top-5 list did not matter.
Crucially, we asked them to twice produce such a list: once after they had looked at the database for only 5 minutes (Speeded Choice; we call the resulting list of products the {\em Speeded Set}), and once after they had studied the database extensively (gold-standard Choice; we call the resulting list of products the {\em gold-standard Set}). Our expectation was that the Speeded and gold-standard Sets produced by participants in the Full Summary condition would be more similar to each other than those of participants in the Baseline Summary condition, because the Full Summary had given them a head start in understanding the database of products.

Each participant was asked to do 2 categories of products. During a pilot experiment, we had observed that some participants had used the `Filter' feature in MS Excel, which had had a very substantial effect on the time used. For the experiment itself we therefore included a question asking participants whether they had used the feature during the experiment. % KvD What did we do with this question? Leave this out maybe?
We also asked participants what information they thought they would want to see added to the summary.
% KvD Did we get any useful responses to this question? Leave this out or say what we learned.
Finally, to ensure that participants would take the experiment seriously, we offered 50 pounds reward for the ``best'' list of products and 20 pounds for the most valuable feed back. -- To summarise the procedure:
% \paragraph{Procedure}
\begin{enumerate}
    \item The participants read the summaries for a minute.
    \item Working with a product database spreadsheet, the participants were given 5 minutes to write their Speeded List on a piece of paper.
    \item With another database spreadsheet, the participants were given another 10 minutes to write their gold-standard Set on another piece of paper.
    \item Afterward the participants answered a questionnaire containing the Likert questions and then key in their preferred product lists on Google Form. % KvD In wh
    \item Then we repeated the procedure for another product category.
\end{enumerate}

\subsubsection{Hypotheses}

Our hypotheses were:

\par Hypothesis 1 [H1]: Likert scores are better for Full Summaries than for Baseline Summaries
\par Hypothesis 2 [H2]: The similarity between Speeded Set and gold-standard Set is greater for participants in the Full Summary conditions than for participants in the Baseline Summary conditions.

For Hypothesis 2, since the sets of products were unordered, we computed similarity between sets of products using the Dice score, a well-known formula for assessing the similarity of sets:
\begin{equation}\label{eq:dice}
Dice(Speeded,GoldStandard) = \frac{2 \times \vert Speeded \cap GoldStandard \vert}{\vert Speeded \vert + \vert GoldStandard \vert}
\end{equation}\index{Dice score}
Here {\em Speeded} is the set of attributes expressed in the description produced by a human author and {\em GoldStandard} is the set of attributes expressed in the Logical Form generated by an algorithm. Dice yields a value between 0 (no agreement) and 1 (perfect agreement).

\subsection{Results}

\subsubsection{Subjective appreciation}

For the Likert part of the experiment, in which we addressed participants' subjective appreciation of the summaries, we found that the participants liked the full summaries better than the baseline summaries in all four respects (i.e., regarding all 4 questions), with statistical significance at p = 0.05 both before and after Bonferroni Correction (Table~\ref{table:Likert Score}).

\begin{table}[ht]
    \caption[Likert Score]{Likert scores of the 4 Questions asked} % title of Table
    \label{table:Likert Score} % is used to refer to this table in the text
    \centering
    \begin{tabular}{lrrrr}
        \Xhline{2\arrayrulewidth}
                                        & Q1     & Q2     & Q3     & Q4     \\
        \Xhline{2\arrayrulewidth}
        N [baseline]                    & 16     & 16     & 16     & 16     \\
        N [exp]                         & 16     & 16     & 16     & 16     \\
        \hline
        mean [baseline]                 & 2.50    & 2.44 & 2.06& 2.69 \\
        mean [exp]                      & 4.00      & 3.69 & 3.69 & 3.88  \\
        SD [baseline]                   & 1.26 & 1.26 & 1.00 & 1.14 \\
        SD [exp]                        & 0.82 & 1.01 & 0.95 & 0.81 \\
        \hline
        p-value (raw)                   & 0.0004 & 0.0043 & 0.0001 & 0.0019 \\
        p-value (Bonferroni Correction) & 0.0016 & 0.0172 & 0.0004 & 0.0076 \\
        \Xhline{2\arrayrulewidth}
    \end{tabular}

    \begin{itemize}
        \item {[baseline]} denotes groups with Baseline Summary
        \item {[exp]} denotes groups with Full Summary
    \end{itemize}
\end{table}

\subsubsection{Task performance}
There was no significant difference, however, between the Dice coefficient from the Baseline Summary (0.1375) and the Full Summary Group (0.1250) (Table~\ref{table:Dice Coeff}); in fact, the Baseline Summaries performed marginally better (not statistically significant) than the Full Summaries.
\begin{table}[ht]
    \caption[Dice Coeff]{Similarity of sets produced in the 2 steps (Dice)} % title of Table
    \label{table:Dice Coeff} % is used to refer this table in the text
    \centering
    \begin{tabular}{lr}
        \Xhline{2\arrayrulewidth}
                        & Dice coefficient \\
        \Xhline{2\arrayrulewidth}
        N [baseline summary]    & 16               \\
        N [full summary]         & 16               \\
        \hline
        mean [baseline] & 0.14           \\
        mean [full summary]      & 0.13           \\
        SD [baseline]   & 0.17           \\
        SD [full summary]        & 0.26           \\
        \hline
        p-value         & 0.8749           \\
        \Xhline{2\arrayrulewidth}
    \end{tabular}
\end{table}

Here, Hypothesis 1 have been confirmed by the result while Hypothesis 2 is inconclusive.
\section{Discussion and Future Work}
\label{sec:discussion}
Results relating to Hypothesis 1 indicate that participants regarded the summaries generated by our algorithm as more useful than Baseline Summaries from the point of view of understanding the product database and selecting a product rapidly and confidently. It seems plausible that the list of common features and price influential features from the summary helped participants to know which columns of the database they should focus their attention on. %Further experiments will need to be carried out to find which part of the summary was really useful and how.

The fact that we were unable to confirm Hypothesis 2 raises interesting questions. We wondered whether a more sophisticated measure of set similarity might lead to a different result, but it turned out that a metric based on cosine similarity (which acknowledges that two products might be different yet share most of their features) did not confirm Hypothesis 2 either.

We considered several possible explanations for the mismatch between subjective appreciation and task performance. Based on the psychology literature on cognitive dissonance reduction \cite{doi:10.1093:scan:nsq054}, one possibility is that participants were reluctant to change the set of product that they had chosen as their Speeded Set. If this was true, one might expect to see gold-standard Sets that were highly similar to the Speeded Sets in both the Baseline Summary condition and the Full Summary condition. However, the low Dice scores (0.14 and 0.13) in Table 2 show that this was not the case.

It is possible that our full summaries appeared useful to participants but that in reality, they were not. Asymmetries of this kind, between perceived and actual usefulness, are surprisingly common; an example is \cite{Law2005} where doctors reported a preference for graphical over textual information presentation although their task performance with textual information was better.

In our view, a more likely explanation is that the setup of our experiment did not do full justice to the idea of Hypothesis 2. In particular, the 10 minutes offered to subjects in the gold-standard choice condition may have been too short. Thus, what we had meant to be a gold standard
%% KK -- the gold-standard got 10 minutes and the speeded choice got 5
may not have been an accurate reflection of participants' real preferences (i.e., not a genuine gold standard). In reality, people would often spend a lot more time buying expensive products with their budget limit in mind.

% It may be that any effects of the summaries were overwhelmed by other factors; for example, it turns out that the average price of the items in our ``gold standard'' lists was significantly lower than that of the items in the speeded lists [ADD NUMBERS], suggesting that subjects used their time to find cheaper items.

In the remainder of this doctoral research, we will explore these issues further.

Following up on the above experiment, and making use of a more detailed study of our corpus of human-written summaries, we have modified our algorithm in a number of ways. First and foremost, we have amplified our summaries to contain a comparison between the products in the target set (e.g., 32-inch TVs) and the products in a natural superset (e.g., TVs). Second, statistical analysis of the corpus has been used to select important features to mention in the summary, and what sentence patterns are employed to talk about them; the patterns involved quantify how frequently a given feature occurs in the target set, for instance ``Most TVs in this category have an HDMI port'', ``Only a few TVs have 4K resolution''. Lastly, a short sentence describing how each importance feature influences prices is included, for example, ``TVs with smart features are more expensive in average''.

In this paper, we have focused on the ability of summaries to provide insight in the content of a set of products. In future research, we want to explore a closely related idea, namely the possibility of using automatically generated summaries of a set of products to {\em explain why} a certain recommendation (i.e., a recommendation for one or more members of the set) is made. The project will address 2 scenarios of set description, which are: descriptions of a predefined set of objects (from databases) in general, and description of a set resulting from a search or a Q/A process.
We also plan to extend the algorithm to generate the list of important features by building and analyzing corpora on recommendations over different product categories available online. When employed in this manner, summaries might be able to boost users' trust in the recommendation and the Recommender System itself.
\pagebreak

% KvD I don't understand what follows, so I've bracketed it out for now:
%% KK - I believe the text below is neither what we want to discuss as our future work for recsys nor what we aim to do for INLG anymore so let's just take them off

% One of the possible way is that we have done analysis on how each feature effects the price. Using that, a product's value can be assigned based on its features. We could use the feature based assigned ``value'' here to further compare the lists produced. Also we came up with a post-hoc hypothesis that consumers would aim for products that are more feature-packed with lower prices. In the future, we plan to extend the algorithm to generate the list of important features by building and analyzing corpora on recommendation over different product categories available online.

% \bibliographystyle{achemso}
\bibliographystyle{plainnat}

\centering
\bibliography{references-kitt,websites-kitt}

\end{document}